\documentclass[letterpaper, 10pt, conference]{ieeeconf}  % Comment this line out if you need a4paper

\IEEEoverridecommandlockouts                              % This command is only needed if 
\usepackage{algorithm}
\let\labelindent\relax
\usepackage{enumitem}
\usepackage{algpseudocode}
\usepackage{tikz}
\usetikzlibrary{automata,calc,positioning,decorations}
\usepackage{graphics} % for pdf, bitmapped graphics files
\usepackage{svg}
\usepackage{hyperref}
\usepackage{booktabs}
\usepackage{multirow}
\usepackage[nolist]{acronym}
\usepackage{cite}
\usepackage{amsmath} % assumes amsmath package installed
\usepackage{amssymb}  % assumes amsmath package installed
\usepackage{subfig}
\usepackage{float}
\usepackage{lipsum}
\usepackage{nth}

\title{\LARGE \bf
Graph-of-Constraints Model Predictive Control for Reactive Multi-agent Task and Motion Planning
}

\author{Anastasios Manganaris, Jeremy Lu, Ahmed H. Qureshi, and Suresh Jagannathan% <-this % stops a space
\thanks{This material is based upon work supported by the Air Force Office of Scientific Research under award number FA9550-24-1-0233. Any opinions, findings, and conclusions or recommendations expressed in this material are those of the author(s) and do not necessarily reflect the views of the United States Air Force.}
\thanks{Anastasios Manganaris, Jeremy Lu, Ahmed H. Qureshi, and Suresh Jagannathan are with
the Department of Computer Science, Purdue University, West
Lafayette, IN, USA, 47907. Email: \texttt{\{amangana, lu1008, ahqureshi\}@purdue.edu} and \texttt{suresh@cs.purdue.edu}}%
}

\begin{document}

\maketitle
% \IEEEpeerreviewmaketitle

\thispagestyle{empty}
\pagestyle{empty}

%%%%%%%%%%%%%%%%%%%%%%%%%%%%%%%%%%%%%%%%%%%%%%%%%%%%%%%%%%%%%%%%%%%%%%%%%%%%%%%%
\begin{abstract}

% Methods for Multi-Agent Task and Motion Planning severly lag behind methods for 

% Main messages:
% existing methods for task and motion planning focus on the single agent case and are limited for the multi-agent case.

% they are limited because they don't support:
% - 
Sequences of interdependent geometric constraints are central to many multi-agent Task and Motion Planning (TAMP) problems. However, existing methods for handling such constraint sequences struggle with partially ordered tasks and dynamic agent assignments. They typically assume static assignments and cannot adapt when disturbances alter task allocations. To overcome these limitations, we introduce \ac{GoC-MPC}, a generalized sequence-of-constraints framework integrated with MPC. GoC-MPC naturally supports partially ordered tasks, dynamic agent coordination, and disturbance recovery. By defining constraints over tracked 3D keypoints, our method robustly solves diverse multi-agent manipulation tasks—coordinating agents and adapting online from visual observations alone, without relying on training data or environment models. Experiments demonstrate that GoC-MPC achieves higher success rates, significantly faster TAMP computation, and shorter overall paths compared to recent baselines, establishing it as an efficient and robust solution for multi-agent manipulation under real-world disturbances. Our supplementary video and code can be found at
\href{https://sites.google.com/view/goc-mpc/home}{https://sites.google.com/view/goc-mpc/home}.

%Formulating long-horizon robotics tasks with sequences of inter-dependent geometric constraints is a promising approach for reactive \ac{TAMP} with several benefits. These include the ability to obtain solutions at real-time speeds, the ability to robustly recover from failures and external disturbances, and no reliance on precise system models or exorbitant data requirements. In multi-agent settings, however, sequences of constraints lack any mechanism for supporting the core requirements of multi-agent \ac{TAMP}, such as supporting partially-ordered tasks and dynamic agent assignments. In this paper, we propose a generalization of these sequence-of-constraint methods to develop a novel algorithm, \ac{GoC-MPC}, which naturally supports these features. We show that \ac{GoC-MPC}, using constraints defined over tracked keypoints in the 3D-scene, can robustly solve a variety of multi-agent manipulation tasks with the ability to coordinate between agents and efficiently recover from disturbances all without any training data or environment models.

% such as the optimization over task-agent assignments, partial ordering of tasks, and multi-agent coordination

% In addition to our \ac{GoC} based reactive task and motion planner, we also contribute a novel partial-conditional-order planning algorithm for generating multi-agent graphs-of-constraints.

\end{abstract}

%%%%%%%%%%%%%%%%%%%%%%%%%%%%%%%%%%%%%%%%%%%%%%%%%%%%%%%%%%%%%%%%%%%%%%%%%%%%%%%%
\section{INTRODUCTION}

% Real world motivation for multi-agent robot teams

The effective deployment of multi-robot teams promises to automate significantly more complex and useful real-world tasks. For example, two robotic arms benefit from a substantially larger effective workspace, the ability to manipulate objects designed for two-handed human use, and the capacity to perform work in parallel. Scaling to larger teams of arms or mobile manipulators further increases throughput and task diversity. However, these capabilities come at the cost of much higher complexity in performing \acf{TAMP}.

% The effective deployment of multi-robot teams holds promise for the ability to automate significantly more complex and useful real-world tasks. Two arms, for instance, benefit from a significantly larger effective workspace, the ability to simultaneously manipulate objects built for two-handed humans, and the ability to do work in parallel. Adding more arms and mobile robots further increases the overall throughput of the robot team, but these increased capabilities come at the cost of significantly higher complexity when performing \acf{TAMP}, especially when considering the possibility of failures and external disturbances.

Despite this complexity, a variety of successful methods for \ac{TAMP} have been developed and remain applicable in the multi-robot setting. A particularly well-studied family of approaches leverages modern constrained optimization to directly solve \ac{TAMP} problems formulated as sequences of constraints \cite{toussaint2015lgp, stouraitis2020onlinehybridmp, stouraitis2020multimodetrajopt}. Just as trajectory optimization underlies online control when used in a receding-horizon loop (e.g., with \ac{MPC}), optimization-based formulations of \ac{TAMP} have also been extended to reactive execution, enabling high-level symbolic plans and low-level continuous motions to be recomputed during runtime \cite{toussaint2022sequenceofconstraintsmpcreactivetimingoptimal}. This line of work demonstrates how re-solving optimization problems in response to new information provides a natural mechanism for handling failures and disturbances. Moreover, recent research has shown that defining constraints in terms of universally applicable geometric primitives---such as spatial keypoints tracked in the robot’s workspace---enables general-purpose manipulation across a wide range of tasks. This allows such methods to achieve impressive generality and avoid the need for precise system models or extensive task-specific training data \cite{huang2024rekepspatiotemporalreasoningrelational}.

\begin{figure} 
\centering

\includegraphics[width=0.9\columnwidth]{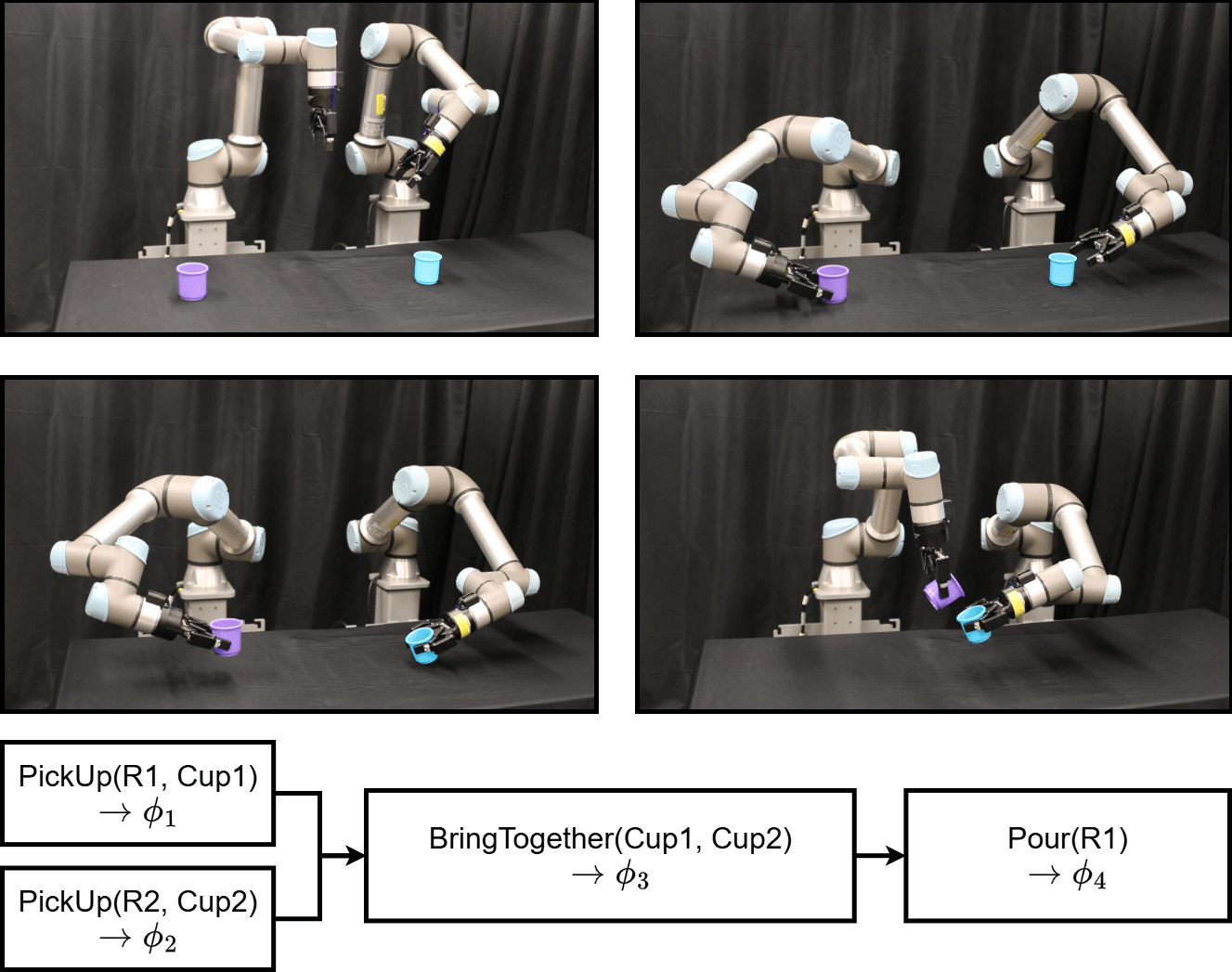}
% \includesvg[width=\columnwidth]{figures/images/intro-showcase.drawio.svg}

\caption{\acfp{GoC} can naturally express partially-ordered multi-agent tasks using a \ac{DAG} of arbitrary system constraints. Our method, \ac{GoC-MPC}, can use a \ac{GoC} with constraints defined in terms of workspace keypoints to find optimal solutions for general multi-agent manipulation tasks using only visual observations. For this two-cup manipulation task, \ac{GoC-MPC} finds an initial solution in 0.373 seconds, and all subsequent solutions in 0.065 seconds on average, demonstrating extremely fast performance.}
\label{fig:showcase}

% This allows for solving multi-agent manipulation tasks fast-enough to react to disturbances that affect the agents at both a symbolic and continuous level

\end{figure}

% Despite this complexity, a variety of successful methods for \ac{TAMP} have been developed that are still applicable to this setting. A particularly well-developed family of approaches have been those leveraging modern capabilities in constrained optimization to solve \ac{TAMP} \cite{toussaint2015lgp, stouraitis2020onlinehybridmp, stouraitis2020multimodetrajopt}. Similar to how trajectory optimization translates to a method for online control by repeatedly formulating and solving an appropriate optimization problem (i.e., with \ac{MPC}), this family of approaches have also been more recently developed as a favorable means for executing \ac{TAMP} in a reactive context \cite{toussaint2022sequenceofconstraintsmpcreactivetimingoptimal}. This allows for both the logical plan of high-level steps and the low-level continuous motions to be continuously updated during execution in order to respond to failures and external disturbances. Subsequent work has also shown that defining these constraints in terms of a universally applicable geometric primitive, such as spatial keypoints tracked within the robot's workspace, can allow for extending the same approach to achieve general manipulation capabilities in many tasks without first obtaining an accurate system model or extensive training data \cite{huang2024rekepspatiotemporalreasoningrelational}. 

% Sequences of constraints as a way to do task and motion planning: Cheap, solvable-at-real-time, doesn't require a complete model, can be constructed from symbolic plans

Although these methods have achieved success in single-agent domains, we identify two fundamental limitations that currently limit their usefulness for general multi-agent tasks. In particular, using singular sequences of constraints imposes a \textbf{total-order} on the steps carried out by the agents and it imposes a \textbf{static} assignment of agents to steps. By artificially imposing ordering constraints between steps that could be otherwise completed in parallel, such as the first two steps in Fig.~\ref{fig:showcase}, the system can unnecessarily halt the progress of one agent when waiting for the other to complete its task. Disturbances can exacerbate this issue arbitrarily, by preventing a single agent from completing an early assigned task, and therefore delaying the entire team, even if they are not affected by the disturbance at all. Static agent assignments can also contribute to this problem, as a poor choice in static assignments without considering the current state of the system can lead to unnecessary extra work or even complete infeasibility. Disturbances should also be able to prompt reassignments of agents if necessary.

% In contrast, a variety of successful approaches exist for \ac{TAMP} in the single-agent setting. A particularly strong trend across these approaches has been optimization, and to leverage modern capabilities of optimization in \ac{TAMP}, a major approach has been formulating \ac{TAMP} problems with a minimal sequence of constraints. This approach was first identified by \cite{toussaint2022sequenceofconstraintsmpcreactivetimingoptimal} as a particularly effective strategy for reactive real-time \ac{TAMP}. Subsequent work also showed that sequences of constraints could alternatively defined in terms of a few tracked spatial keypoints in a robot's workspace \cite{huang2024rekepspatiotemporalreasoningrelational}. 

% Limitations of sequences of constraints for multi-agent scenarios

In this work, we address these limitations by realizing the generalization of Sequences-of-Constraints to \emph{Graphs}-of-Constraints (GoCs). \acp{GoC} are \acp{DAG} of constraints derived from symbolic, high-level actions, and therefore can naturally represent \textbf{partially-ordered} sets of task steps. Furthermore, constraints within a \ac{GoC} are defined with respect to a \textbf{dynamic} assignment of agents to subtasks, which allows for optimizing over this assignment when obtaining the solution. To solve problems formulated with \acp{GoC}, we also contribute \ac{GoC-MPC}, which decomposes and solves the optimization problem resulting from a \ac{GoC} at real-time speeds, therefore enabling reactive multi-agent \ac{TAMP}. We evaluate our algorithm on several bimanual manipulation tasks that take advantage of properties of the \ac{GoC} formulation, including block stacking (featuring parallelized pick-ups and sequenced placements), transferring liquids in cups (featuring coordination between agents based on the objects they are manipulating), and tablecloth folding (featuring exact synchronization of agents).

% Our approach overall demonstrates efficacy in

\section{RELATED WORK}

In general, the field of \ac{TAMP} is concerned with extending symbolic, discrete planning algorithms \cite{fikes1971strips, weld1994leastcommitmentplanning, jorg2001ff} with continuous motion planning \cite{lavalle06planningalgs, qureshi2019mpnets} to solve complex, long-horizon robotics tasks \cite{garrett2020pddlstream, vu2024coast}. Approaches to \ac{TAMP} include constraint-based \cite{dantam2018idtmp}, sampling-based \cite{garrett2020pddlstream}, and the aforementioned optimization-based methods \cite{zhao2025tampsurvey}. The latter \ac{TAMP} methods formulate a continuous or mixed-integer optimization problem \cite{toussaint2015lgp, stouraitis2020onlinehybridmp, stouraitis2020multimodetrajopt, zhao2025tampsurvey} and therefore benefit from naturally being able to incorporate diverse costs and constraints. Optimization-based approaches have also been extended to multi-agent settings \cite{antonyshyn2023multimobilerobottampsurvey}.

Reactive \ac{TAMP} is a closely related area to \ac{TAMP} that, instead of assuming that \ac{TAMP} solutions can be perfectly executed without failure or external disturbances, focuses on the ability to quickly recompute plans online during execution. Specific strategies include synthesizing hierarchical policies for symbolic actions via reactive synthesis from temporal logic specifications \cite{livingston2012backtracking, guo2013modelchecking, he2019reactivesynthesis}, defining object-relative plans that remain valid despite object motion \cite{migimatsu2020object}, and re-solving optimization problems derived from symbolic skeleton plans \cite{braun2022rhhlgprecedinghorizonheuristicsbased, xue2024dlgp, zhang2024reactivetampmppi, toussaint2022sequenceofconstraintsmpcreactivetimingoptimal}. To our knowledge, we contribute the first multi-agent-capable reactive \ac{TAMP} method belonging to this latter category.

% We note here that these latter methods do not compute entirely new task level plans, as in \cite{paxton2019logicaldynamicalsystems, li2021reactivetampltl}, but operate with respect to one task skeleton. The controller then completes the steps in this task skeleton and, in the case of disturbances, ``backtracks'' and recompletes prior steps as necessary.

We also note that this last class of solutions that reactively optimize with respect to a task skeleton has received increased attention as a way to relax the heavy modeling assumptions of traditional \ac{TAMP}.
In particular, \acp{ReKep} \cite{huang2024rekepspatiotemporalreasoningrelational} are constraints on the geometric relationships of keypoints within an environment, which enables the execution of general manipulation skills without requiring exhaustive world modeling or large, task-specific datasets as with other methods \cite{manuelli2022kpam, reed2022a, chi2023diffusionpolicy}.
Our method is also designed to use this constraint representation, and therefore achieves similar generality, while still addressing the gaps in prior reactive \ac{TAMP} methods for supporting dynamic agent assignment and parallelizable task structures.

\section{METHODS}

In this section, we present our approach for achieving reactive multi-agent \ac{TAMP}. We begin by discussing the preliminaries for our method, including necessary notations and the limitations of prior methods in supporting multi-agent tasks (Section~\ref{sec:preliminaries}). We then describe \acp{GoC} and how they specifically address these limitations, as well as the optimal control problem for \ac{TAMP} that results from a \ac{GoC} (Section~\ref{sec:graphs-of-constraints}). Next, we describe the specific steps that are necessary to decompose and efficiently solve this problem within a receding-horizon \ac{MPC} scheme to enable reactivity in multi-agent \ac{TAMP} (Section~\ref{sec:problem-decomposition}). The final algorithm, \ac{GoC-MPC}, is summarized in Section~\ref{sec:complete-algorithm}.

\subsection{Preliminaries}
\label{sec:preliminaries}

% We then discuss the issues that arise when applying these sequences to the multi-agent domain. Finally, we formally define \acp{GoC} and the resulting optimal control problem with which one can efficiently perform reactive long-horizon multi-agent manipulation.

As mentioned above, our method is based on the idea that robotic tasks can be defined in terms of ordered constraints on the robotic system state, inspired by methods such as \cite{toussaint2022sequenceofconstraintsmpcreactivetimingoptimal, huang2024rekepspatiotemporalreasoningrelational}. Let $X \subset \mathbb{R}^d$ denote the $d$-dimensional configuration space for a robot system. The configuration space $X = X_a \times X_p$ consists of an actuated component $X_a \subset \mathbb{R}^{d_a}$ for the controllable robot configurations and a non-actuated passive component $X_p \subset \mathbb{R}^{d_p}$ for the remaining scene elements (e.g., keypoints on objects). In a multi-agent system, the actuated configuration space $X_a = X_{a_1} \times X_{a_2} \times \ldots \times X_{a_M}$ can be split into $M$ single-agent configuration spaces $X_{a_j}$ for $j \in [1, M]$. For any configuration space $X$, we denote its tangent space as $\dot{X}$

Constraints are functions denoted by $\phi_i : X \rightarrow \mathbb{R}^{d_i}$ with $d_i \in \mathbb{N}$ where $\phi_i(x) \leq 0$ indicates satisfaction of the constraint under configuration $x$. A sequence of constraints, as defined in \cite{toussaint2022sequenceofconstraintsmpcreactivetimingoptimal, huang2024rekepspatiotemporalreasoningrelational}, is then a pair  $(\phi_{1:K}, \bar{\phi}_{1:K})$ of $K$ ``waypoint'' constraint functions and $K$ ``path'' constraint functions. Each waypoint is associated with a specific time $t_i$ which satisfies $t_{i-1} \leq t_i$. Under this formulation, the objective is then to find an optimal path $\xi : [0, t_K] \to X$ that satisfies every constraint at the appropriate times.

\textbf{Limitation of existing constraint representations.} As mentioned above, the property that $\forall i \in [1, K] \; t_{i-1} \leq t_i$ requires that the steps in the trajectory $\xi$ are totally-ordered. Therefore, it is impossible to specify stages that should be done in parallel or transposed as necessary for optimality. Furthermore, this formulation implies that multiple agents are always implicitly treated as a single ``joint-agent'', and every constraint $\phi$ must either operate over the joint-agent configuration  or a static subset of that joint-agent configuration. We consider this static assignment issue the second major issue for the existing formulation.

\subsection{Graphs of Constraints}
\label{sec:graphs-of-constraints}

We now define \acp{GoC}, which are a generalization of the sequences-of-constraints used in prior work and immediately address the limitations of a totally-ordered sequence and statically assigned constraints. For the former, we instead define a partial order for these constraints, which naturally results in defining a \ac{DAG} structure. For the latter, we extend the definition of constraints to consider a dynamic set of assignments between $M$ agents and $K$ assignable ``subtasks''. We can represent these assignments with a matrix $A \in \{0, 1\}^{K \times M}$ of binary decision variables with $K$ rows and $M$ columns. This matrix is constrained to be row stochastic to ensure that only a single agent is assigned to each of the $K$ subtasks. A constraint function is now defined to be a function $\phi_i : \{0, 1\}^{K \times M} \times X \to \mathbb{R}^{d_i}$ where $\phi(A, x) \leq 0$ indicates satisfaction of the constraint under assignment $A$ and configuration $x$. The additional input allows now for the definition of \emph{disjunctive} constraints that apply specifically to the agents selected for the relevant subtasks in $A$. In our implementation, these disjunctive constraints were implemented using big-$M$ gating \cite{grossmannruiz2012gdp}.

Let $\Phi$ denote the set of all such constraint functions. Then, a \ac{GoC} is the tuple
\[
\mathcal{G} = (V, E, \Phi_V, \Phi_E),
\]
where $V \subset \mathbb{N}$ is a finite set of $N$ nodes (i.e., step indices), $E \subseteq V \times V$ is a set of directed edges, $\Phi_V : V \to \Phi$ is a mapping from the nodes to ``waypoint'' constraint functions, and $\Phi_E : E \to \Phi$ is a mapping from the edges to ``edge'' constraint functions. For brevity, we denote $\Phi_V(v)$ as $\phi_v$ and $\Phi_E(e)$, where $e = (a, b)$, as $\bar{\phi}_e$ or $\bar{\phi}_{ab}$.

Given a \ac{GoC}, the objective is then to find a set of agent assignments $A$, an optimal and constraint-satisfying system trajectory $\xi : [0, t_{\max}] \rightarrow X$, and a set of timings $t_{1:N}$ for each node in the \ac{GoC}. The trajectory is defined over a time interval ending at the maximum time needed to finish the task $t_{\max} = \max_{1\leq i \leq N} t_i$ and is required to satisfy every waypoint and edge constraint in that interval at the appropriate times. Furthermore, we typically are interested in solving the task in the minimum amount of time and with a minimal amount of other costs. These constraints and this objective then can be expressed together as
\begin{align}
    \label{eqn:full-objective}
    \min_{A, \xi, t_{1:N}} &\quad t_{\max} + \int_0^{t_{\max}} c(\xi(t), \dot{\xi}(t), \ddot{\xi}(t)) \; dt \\
    \text{s.t.} &\quad \xi(0) = x_0 \quad \dot\xi(0) = \dot{x}_0 \quad t_0 = 0 \nonumber \\
    &\quad \forall k \in \{0, 1, \ldots,K\} : \sum_{j = 0}^{M} A(k, j) = 1 \nonumber \\
    &\quad \forall {v \in V} : \phi_v(A, \xi(t_v)) \leq 0  \nonumber \\
    \begin{split}
        &\quad \forall (a,b) \in E : t_a \leq t_{b} \; \wedge \\
        &\quad \quad \quad \forall t \in [t_{a}, t_b] : \bar{\phi}_{ab}(A, \xi(t)) \leq 0  \nonumber
    \end{split} \\
    \label{eqn:dynamical-constraint}
    &\quad \forall t : \dot \xi(t) = h(\xi(t)),
\end{align}
where $c$ represents an arbitrary cost function that can penalize acceleration or collisions. The solution trajectory is constrained to originate at the current system state $x_0$ with the current velocity $\dot{x}_0$, and the task assignments $A$ are constrained such that only one agent is assigned to a single task. The final constraint \eqref{eqn:dynamical-constraint} represents the dynamical constraints of the system. For instance, the movement of a cube is constrained to match the movement of the robot assigned to hold that cube and is otherwise affected by gravity.

% The remaining constraints constrain the timing for each achieved waypoint to be ordered according to the edges in $\mathcal{G}$ and the states at those times must satisfy the appropriate waypoint constraints $\phi_v$ and edge constraints $\phi_e$. 

% \subsection{Constraint-Graph Generation}
% Constraints are generated from a plan.

% \subsection{Partial-Conditional-Ordering Planner}
% Novel planner is necessary.

\subsection{Problem-Decomposition}
\label{sec:problem-decomposition}

As with other methods leveraging optimization for long-horizon tasks, the full optimization problem is difficult to solve and impractical for a fast reactive planner. Instead, the problem can be decomposed and a satisfactory approximation can be obtained by solving for the three resulting components: (1) the waypoints and agent assignments, (2) the trajectory timings and velocities, and (3) a short horizon path for all agents. We describe this decomposition here and how each subproblem is extracted from a \ac{GoC}.

\paragraph{Waypoints and Assignments Sub-Problem}
\label{sec:wp-and-assignment-problem}

The first component of the decomposed problem solves for an optimal set of waypoints, represented by a vector $W \in X^N$ of $N$ configurations, along with the assignments of agents to tasks, $A$. Let $W(v)$ for $v \in V$ denote the waypoint at node $v$, and let $V_\text{next}$ denote the topologically-first set of nodes. To encourage minimizing the total time as in \eqref{eqn:full-objective}, we use a surrogate objective based on the total geodesic distance $\bar{c} : X \times X \to \mathbb{R}$ between the waypoints on every edge of a \ac{GoC} and between the initial system state $x_0$ to the nodes in $V_\text{next}$:
\begin{align*}
    \min_{W, A} &\quad \sum_{v \in V_\text{next}} \bar{c}(x_0, W(v)) + \sum_{a,b \in E} \bar{c}(W(a), W(b)) \\
    \text{s.t.} &\quad \forall k \in \{1, \ldots,K\}  : \sum_{j = 1}^{M} A(k, j) = 1 \\
    &\quad \forall {v \in V} : \phi_v(A, W(v)) \leq 0 \\
    &\quad \forall {(a, b) \in E} : \bar{\phi}_{ab}(A, W(a)) \leq 0 \wedge \bar{\phi}_{ab}(A, W(b)) \leq 0 \\
    &\quad \forall {(a, b) \in E} : \bar{h}(A, W(a), W(b)) \leq 0.
\end{align*}
The remaining constraints adapt the original constraints for enforcing the assignment matrix is row-stochastic, every waypoint constraint is respected, and every edge constraint is respected. The joint optimization of waypoints and agents allows for resolving the subset of constraints that are dependent on agent assignments.

The final constraint, defined by $\bar{h}$, is meant to approximate the consequences of the dynamical constraints \eqref{eqn:dynamical-constraint} in the transitions between waypoints. In our implementation, we reuse the information provided by the edge constraints $\Phi_E$ to determine across which edges are keypoints manipulated. Using this information, we define $\bar{h}$ by making two assumptions: (i) points not being manipulated by a robot remain fixed, and (ii) points being manipulated move rigidly with the robot and with each other. Note that these constraints are also dynamically gated depending on the agent assignments in $A$.

% Note that this doesn't conflict with the ability for the human to disturb the system as those disturbances are reflected in $X(0)$. Furthermore, this constraint allows for complex reasoning far into the future about the position of objects. For instance, this constraint ensures that every waypoint topologically following a robot moving block A onto block B will consider block A's new position, and if block C is moved onto block A, then the appropriate location of block C is automatically deduced. 

The resulting problem is a mixed-integer non-linear program, but is small enough to be efficiently, locally solved with enumerating the possible integer variable solutions and solving for each assignment the non-linear program using Ipopt \cite{wachter2006ipopt}. On the first instantiation of this problem, we initialize the waypoint decision variables $W(v)$ for all $v \in V$ with the initial state $x_0$, and on all subsequent iterations they are initialized with the previous iteration's solution.

\begin{figure}
\centering

    \begin{tikzpicture}[
      >=latex, node distance=14mm, thick,
      state/.style={draw, circle, minimum size=8mm, inner sep=0pt},
      ghost/.style={draw, circle, minimum size=8mm, inner sep=0pt, dashed}
    ]

    % dashed “prior” states on the left
    \node[ghost] (g0) {$t_0$};
    \node[ghost, right=12mm of g0, yshift=1.0cm] (g1) {$t_1$};

    %--- column of main states (single column)
    \node[state, right=12mm of g0, yshift=-1.0cm] (s2) {$t_2$};
    \node[state, right=12mm of g1] (s3) {$t_3$};
    \node[state, right=12mm of s2] (s4) {$t_4$};
    \node[state, right=12mm of s3, yshift=-1.0cm] (s5) {$t_5$};

    % dashed links from ghost states
    \draw[dashed,->] (g0) -- (g1);
    
    % main edges
    \draw[->] (s3) -- node[right] {$\leq$} (s4);
    
    \draw[->] (s3) -- (s5);
    \draw[->] (s4) -- (s5);
    \draw[->] (s2) -- (s4);
    \draw[->] (g1) -- (s3);
    \draw[->] (g0) -- (s2);
    
    % --- offsets for the two Δ arcs
    \coordinate (s3top) at ($(s3.north)+(-10mm,2mm)$);
    \coordinate (s5top) at ($(s5.north)+(0,2mm)$);
    
    \coordinate (s2bot) at ($(s2.south)+(-10mm,-0mm)$);
    \coordinate (s5bot) at ($(s5.south)+(0,-4mm)$);
    
    % Δ_1 (top) – starts 6mm above s3, ends 6mm above s5
    \draw[->,shorten <=2pt,shorten >=2pt]
      (s3top) .. controls ($(s3top)+(10mm,0mm)$) and ($(s5top)+(-10mm,10mm)$)
      .. node[pos=.2, above] {$\Delta_1(0)$}
         node[pos=.8, above, yshift=2mm] {$\Delta_1(1)$}
         node[pos=.95, right] {$\xi_1$} (s5top);
    
    % Δ_2 (bottom) – starts 6mm below s2, ends 6mm below s5
    \draw[->,shorten <=2pt,shorten >=2pt]
      (s2bot) .. controls ($(s2bot)+(10mm,-5mm)$) and ($(s5bot)+(-8mm,-8mm)$)
      .. node[pos=.15, below] {$\Delta_2(0)$}
         node[pos=.52, below, yshift=-1mm] {$\Delta_2(1)$}
         node[pos=.95, below, yshift=-5mm] {$\Delta_2(2)$}
         node[pos=.95, right] {$\xi_2$} (s5bot);
     
    % % big brace for R under the column
    % \draw[decorate, decoration={brace, amplitude=6pt}]
    %   ($(s5.south west)+( -8mm,-10mm)$) -- ($(s2.north west)+(-8mm, 10mm)$)
    %   node[midway, below=9pt, align=center] {$R$};
    
    \end{tikzpicture}

\caption{After waypoints $W$ and agent assignments $A$ have been determined, splines for each agent are threaded through the appropriate waypoints in order. The current subgraph is determined by a subset of remaining nodes $R$ (denoted with solid borders) that shrinks as tasks are completed (denoted with dashed borders). The graph's structure determines additional constraints that can enforce ordering or ensure synchronization between agents' actions.}
\label{fig:goc-visualization}

\end{figure}
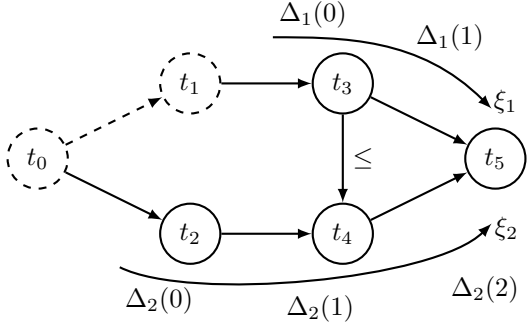

\paragraph{Agent Splines Problem}
\label{sec:timing-problem}

% Therefore, the decision variables for the resulting 
% $W_j$, the cubic spline can be defined with another set of configuration velocities 

After solving for the waypoints and agent assignments, we solve for an optimal multi-agent path passing through them. We represent this multi-agent path with a set of agent-specific cubic splines $\{\xi_1, \ldots \xi_M\}$ collectively starting from $x_0$ with velocity $\dot{x}_0$ and passing through the appropriately ordered agent-specific waypoints $W_j \in X_{a_j}^{N_j}$ for $j \in [1, M]$. The number of waypoints relevant to agent $j$ is denoted by $N_j$. Each spline also passes through its $i$\textsuperscript{th} waypoint with a velocity denoted by $V_j(i) \in \dot{X}_{a_j}$. To avoid the possibility of invalid timings, the timing of the spline is determined by a vector of time-deltas $\Delta_j \in \mathbb{R}^{N_j+1}$, where $\Delta_j(i)$ gives the time for agent $j$ to travel between the $i$\textsuperscript{th} and the $i+1$\textsuperscript{th} node on their spline. Based on these values, it is convenient to compute dynamical costs for the entire spline, denoted by $c(W_j, V_j, \Delta_j)$, including its total acceleration, max velocity, max acceleration, and max jerk. An illustration of the overall concept of splines threaded through a \ac{GoC} is provided in Fig.~\ref{fig:goc-visualization}. 

We identify the ordered waypoints for each agent with a topologically-ordered breadth-first traversal of the \ac{GoC}. For each traversed node $v$, agent $j$'s component of the waypoint $W(v)$ is added to $W_j$ for every agent $j$ that is relevant to the constraints $\phi_v$. Relevance is determined by either the static definition of $\phi_v$ or dynamic assignments in $A$. During this traversal, additional inter-agent timing constraints can be deduced. A set of ``less-than'' constraints $E_{\leq}$ is obtained from the traversal's non-tree edges $(a, b)$, enforcing the relevant order of two agent's waypoints. Each edge is recorded as a tuple $(j_a, l_a, j_b, l_b)$ where $j_a, j_b$ are two agents that visit node $a$ and $b$, and $l_a, l_b$ are the indices of $a$ and $b$ in their respective trajectories. For nodes in which two trajectories converge, a similar set $E_{=}$ of tuples $(j_a, l_a, j_b, l_b)$ can be obtained to enforce the dynamic synchronization of a pair of agents. For example, the edge from node 3 to node 4 in Fig.~\ref{fig:goc-visualization} forces agent 1 to reach its configuration in waypoint 3 before agent 2 reaches its configuration in waypoint 4. The convergence of both splines at node 5 require that agent 1 and agent 2 both arrive reach their configurations in waypoint 5 simultaneously. These constraints are represented by tuples $(1, 0, 2, 1) \in E_\leq$ and $(1, 1, 2, 2) \in E_{=}$ respectively.

In general, the agent spline subproblem is
\begin{align*}
    \min_{V_{1:m}, \Delta_{1:m}} &\quad \sum_{j=1}^{M} \sum_{i = 0}^{N_j} \Delta_j(i) + c(W_j, V_j, \Delta_j) \\
    \text{s.t.} &\quad \forall j \in \{1, 2 \ldots M \} : V_j(0) = \dot{x}_{0_j} \wedge V_j(N_j) = 0 \\
    &\quad \forall {(j_a,l_a, j_b,l_b) \in E_{\leq}} : \sum_{i=0}^{l_a} \Delta_{j_a}(i) \leq \sum_{i=0}^{l_b} \Delta_{j_b}(i) \\
    &\quad \forall {(j_a,l_a, j_b,l_b) \in E_{=}} : \sum_{i=0}^{l_a} \Delta_{j_a}(i) = \sum_{i=0}^{l_b} \Delta_{j_b}(i),
    % &\quad \dot{\xi}(0) = 0, \; \dot{\xi}(t_K) = 0 \\
\end{align*}
which is a quadratic program that can be efficiently solved with the optimization routines implemented in MOSEK \cite{mosek2025}.

% the first instantiation of this problem, we initialize the waypoint decision variables $W(v)$ for all $v \in V$ with the initial state $x_0$, and on all subsequent iterations they are initialized with the previous iterations solution.

% If there is order ambiguity, solve for a specific order. This is technically a traveling salesman problem, but it is quite small and can be brute-forced or approximated if necessary. This is minor.

\paragraph{Short Receding-Horizon Problem}
\label{sec:short-receding-horizon-problem}

Finally, using the smooth reference trajectories computed in the last subproblem, aggregated into a single joint-agent trajectory $\xi^*(t) = \langle \xi_1(t), \xi_2(t), \dots \xi_M(t)\rangle$, we can compute a short receding horizon trajectory of $H$ steps spaced out by a time interval $dt$ that tracks the reference trajectory as close as possible while also taking into account more fine dynamical costs, represented by $c_h$, such as environment collision costs and reachability costs. This results in the objective
\begin{align*}
    \min_{\xi} &\quad \int_{0}^{H} || \xi(t) - \xi^*(t)||^2 + c_h(\xi(t)) \\
    \text{s.t.} &\quad \xi(0) = x,
\end{align*}
which is again a quadratic program that is efficiently solved with MOSEK \cite{mosek2025}.

\begin{algorithm}[t]
\small
\caption{GoC-MPC Cycle} \label{alg:goc_mpc_cycle}
\begin{algorithmic}[1]
    \Require A \ac{GoC} $\mathcal{G}$, the remaining nodes $R \subseteq V$, time-delta cutoff $\tau$, constraint tolerance $\epsilon$, current state and velocity $x$, $\dot{x}$
    % \State $S \leftarrow \Call{GetNextStages}{\mathcal{G}_s}$
    \For{$(a,b) \gets \Call{CutEdges}{\mathcal{G}, R}$}    \Comment{$E \cap ((V \setminus R) \times R)$}
        \If{$\bar{\phi}_{ab} \geq \epsilon$}
            \State $R \leftarrow R \cup a$ \Comment{Phase Backtracking}
            \State \Return
        \EndIf
    \EndFor
    \State $\mathcal{G}_s \leftarrow \Call{Subgraph}{\mathcal{G}, R}$
    \State $P_{1} \leftarrow \Call{ConstructWaypointProblem}{\mathcal{G}_s, x}$
    \State $W, A \leftarrow \Call{Solve}{P_{1}}$
    \Comment{Sec. III a.}
    \State $\{W_j\}_{j=1}^{M} \leftarrow \Call{GetAgentPaths}{\mathcal{G}_s, W, A}$
    \State $P_{2} \leftarrow \Call{ConstructTimingProblem}{\{W_j\}_{j=1}^{M}, x, \dot{x}}$
    \State $\{\Delta_j, V_j\}_{j=1}^{M} \leftarrow \Call{Solve}{P_{2}}$    \Comment{Sec. III b.}
    \For{$j \gets 1$ to $M$}
        \State $v_j \leftarrow \Call{GetNextNode}{j}$
        \If{$\Delta_j(0) \leq \tau$}
            \If{$\phi_{v_j} \leq \epsilon$}
                \State $R \leftarrow R \setminus v$ \Comment{Phase Progression}
                % \State $\xi_j \leftarrow \Call{CubicSpline}{X_j, V_j, \Delta_j}$
            \EndIf
        \EndIf
    \EndFor
    \For{$j \gets 1$ to $M$}
        \State $\xi_j \leftarrow \Call{CubicSpline}{W_j, V_j, \Delta_j}$
    \EndFor
    \State $P_{3} \leftarrow \Call{ConstructShortPathProblem}{\{\xi_j\}_{j=1}^{M}, x}$
    \State $\xi_h \leftarrow \Call{Solve}{P_{3}}$
    \Comment{Sec. III c.}
    \State \Return $\xi_h$
\end{algorithmic}
\end{algorithm}

\subsection{Graph-of-Constraints MPC}
\label{sec:complete-algorithm}

In this section, we describe the complete algorithm, shown in Algorithm~\ref{alg:goc_mpc_cycle}, that iteratively solves the above three subproblems throughout execution in order to perform reactive \ac{TAMP}. Apart from these optimization problems, the algorithm generalizes the phase progression and backtracking components of prior work \cite{toussaint2022sequenceofconstraintsmpcreactivetimingoptimal, huang2024rekepspatiotemporalreasoningrelational} so that progression occurs topologically forward or backward through the \ac{GoC} and is monitored through a set of nodes $R \subseteq V$ that represent the remaining steps in the task.

To implement forward phase progression, we examine the computed time at which each agent is expected to cross the next node, which is given in $\Delta_j(0)$, and compare it to a small time-delta cutoff value $\tau$. If the crossing is expected to occur within the cutoff period, all constraints associated with that node are checked and, if they are satisfied, the node is removed from the set of remaining nodes. For backtracking, one instead examines the edge constraints for each edge crossing between one of the remaining nodes in $R$ and the rest of the completed graph. Here, if any edge constraint is violated, the source node associated with that edge is added back into $R$ and the cycle restarts. This makes it so that repeated cycles will backtrack until edge constraints are satisfied. Finally, the set of remaining nodes is used in each cycle to obtain a subgraph $\mathcal{G}_s$ from the overall \ac{GoC} $\mathcal{G}$ that determines the structure of the waypoint optimization problem and of the splines relevant to each agent. As a result, the entire controller allows for flexible agent parallelization, progression through tasks, and recovery from failures.

\section{EXPERIMENTS}
%In this section, we describe the experiments conducted to validate GoC-MPC as an efficient and robust method for reactive \ac{TAMP}. We first compared our method's computational scalability with ReKep \cite{huang2024rekepspatiotemporalreasoningrelational}, which is the reactive \ac{TAMP} method most comparable to ours.
%We then separately evaluate our methods improved ability to robustly handle disturbances in a multi-agent setting. Finally, we examine our method's effectiveness in completing the following real-world tasks: stacking objects, pouring liquids, and folding garments.

\begin{figure*}[h]
\centering

% \includegraphics[width=0.8\textwidth, height=0.5\textwidth]{example-image}

% \includesvg[width=\textwidth]{figures/images/experiment-settings.drawio.svg}

\subfloat[Pick-and-Pour]{%
   % \includesvg[width=0.30\linewidth]{figures/compressed_images/pick-and-pour-task.drawio_compressed.svg}
    \includegraphics[width=0.30\linewidth]{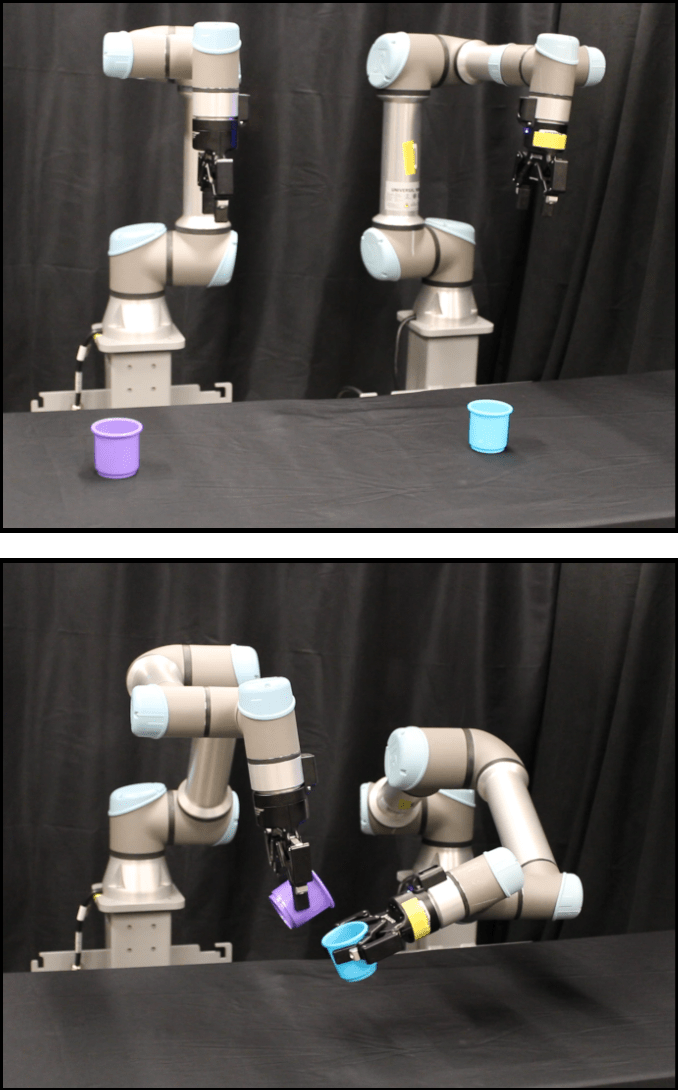}
   \label{fig:pick-and-pour}
}
% \hspace{-12pt}
\subfloat[Block Stacking]{
    % \includesvg[width=0.30\linewidth]{figures/compressed_images/block-stacking.drawio_compressed.svg}
    \includegraphics[width=0.30\linewidth]{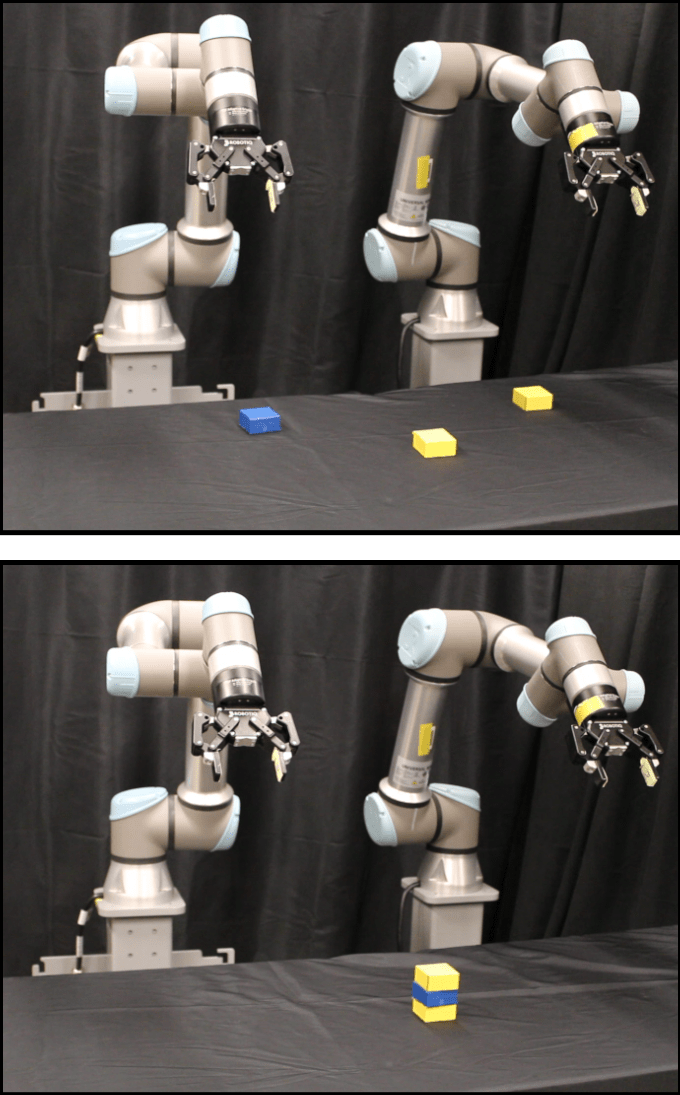}
    \label{fig:block-stacking}
}
\hspace{0.8pt}
\subfloat[Tablecloth Folding]{%
    % \includesvg[width=0.30\linewidth]{figures/compressed_images/folding.drawio_compressed.svg}
    \includegraphics[width=0.30\linewidth]{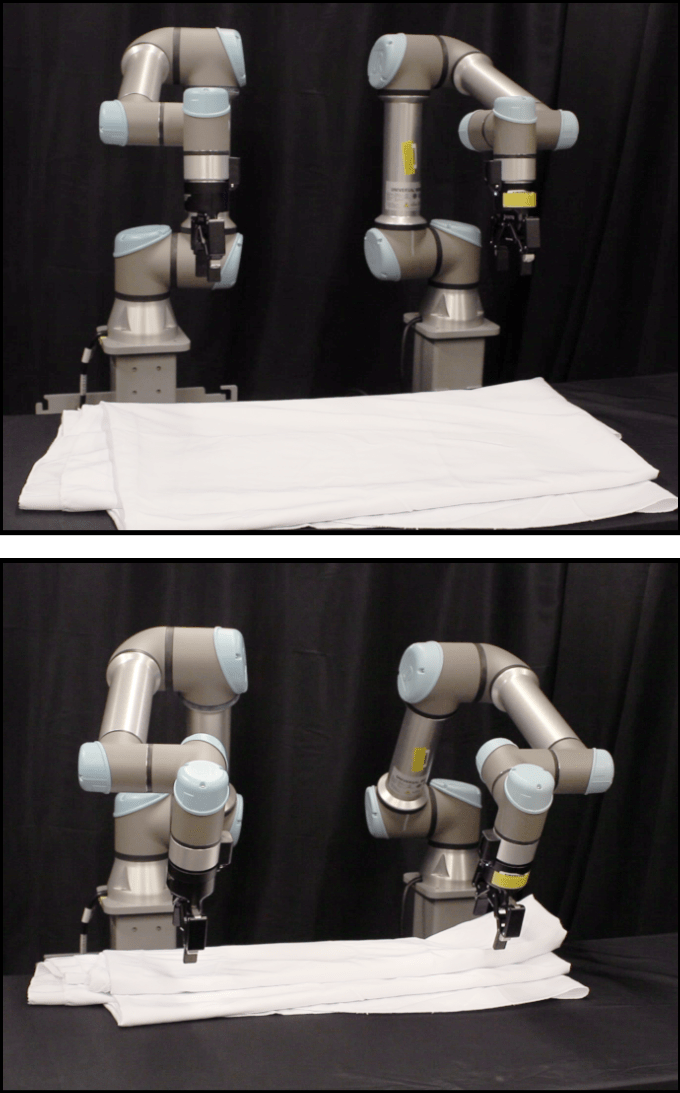}
    \label{fig:folding}
}

\caption{We conducted experiments in two simulated and three real bimanual maniulation tasks with two UR5e arm robots. These included a distant-cup pick-and-pour task, a three-block stacking stacking task, and a tablecloth folding task (not simulated due to including a soft-body). These tasks demonstrate that our method is capable of handling partially-ordered tasks, coordination of multiple robots based on the objects they are manipulating, and synchronization between agents.}

\label{fig:experiment-settings}

\end{figure*}

In this section, we describe the experiments conducted to validate GoC-MPC as an efficient and robust method for reactive \ac{TAMP}. We evaluate the approach on three representative tasks:

\begin{itemize}[leftmargin=*]
    \item \textbf{Block-Stacking}: The robot begins with three colored blocks positioned flat on the workspace. The task requires grasping and stacking each block to construct a stable
    three-block tower.
    \item \textbf{Pick-and-Pour}: The system is initialized with a pouring cup and empty target container, both in upright positions. The robot must transfer the liquid between containers while maintaining precise control to prevent spillage.
    \item \textbf{Cloth-Folding}: The manipulation task involves a rectangular cloth initially laid flat on the work surface. The robot executes a series of grasping and folding motions to bring the cloth's corner regions inward, achieving a compact, organized configuration.
\end{itemize}
The visualization of these tasks is shown in Fig. \ref{fig:experiment-settings}. 

\noindent\textbf{Baselines.} While several methods address TAMP tasks, they typically require complete world models and cannot operate directly from visual observations. To the best of our knowledge, ReKep \cite{huang2024rekepspatiotemporalreasoningrelational} is the only method that solves reactive TAMP tasks directly from keypoints extracted from visual input, without relying on explicit world models. We therefore select ReKep as our primary baseline.

\noindent\textbf{Experimental analyses.} We conducted four sets of evaluations:
\begin{itemize}[leftmargin=*]
    \item \textbf{Static setting (without external disturbance):} Comparison against baselines in block-stacking and pick-and-pour tasks using visual observations (Section ~\ref{sec:staticanalysis}).
    \item \textbf{Disturbance setting:} Comparison against baselines on block-stacking tasks under external disturbances, where agents are explicitly disrupted (e.g., objects pushed from robot hands or object locations altered), requiring reactive TAMP to reobserve changes and dynamically adapt task assignments and motion plans (Section ~\ref{sec:disturbancesanalysis}).
    \item \textbf{Scalability:} Analysis of performance against the number of agents and objects in block-stacking tasks (Section ~\ref{sec:scalabilityanalysis}).
    \item \textbf{Real-world validation:} Demonstration of our approach on all three tasks in physical settings (Section ~\ref{sec:realworldanalysis}).
\end{itemize}

\noindent\textbf{Metrics.} To quantify performance, we evaluate results using:
\begin{itemize}[leftmargin=*]
    \item \textbf{Success:} Proportion of trials completed successfully.
    \item \textbf{Max Time (s):} Maximum wall-clock execution time across all MPC cycles.
    \item \textbf{Average Time (s):} Average wall-clock execution time per MPC cycle.
\end{itemize}

\begin{table}[t]
\centering
    %\scriptsize
    %\setlength{\tabcolsep}{4.0pt}
    \caption{Comparison of our approach with ReKep on the \textbf{Block-Stacking} task shows that our method outperforms across all metrics. It achieves a 100\% success rate and, on average, runs 70 $\times$ faster than ReKep, highlighting its effectiveness for reactive TAMP.}
    \label{tab:blocks}
    \begin{tabular}{lcc}
        \toprule
        & \textbf{GoC-MPC (Ours)} & \textbf{ReKep} \\
        \midrule
        Success Rate & $\boldsymbol{10/10}$ & $7/10$ \\
        %\midrule
        Max Time (s) & $\boldsymbol{0.520 \pm 0.311}$ & $8.38 \pm 0.062$ \\
        %\midrule
        Avg. Time (s) & $\boldsymbol{0.108 \pm 0.012}$ & $7.11 \pm 0.186$ \\
        %\midrule
        Total Length (m) & $\boldsymbol{2.54 \pm 0.216}$ & $4.75 \pm 2.59$ \\
        \bottomrule
    \end{tabular}
\end{table}

\begin{table}[t]
\centering
    %\scriptsize
    % \setlength{\tabcolsep}{4.0pt}
    \caption{Comparison of our approach with ReKep on the \textbf{Pick-and-Pour} task. Our method achieves a 100\% success rate, runs 40 $\times$ faster than ReKep, and produces significantly shorter path lengths than the baseline.
    }
    \label{tab:pickpour}
    \begin{tabular}{lcc}
        \toprule
        & \textbf{GoC-MPC (Ours)} & \textbf{ReKep} \\
        \midrule
        Success Rate & $\boldsymbol{10/10}$ & $6/10$ \\
        %\midrule
        Max Time (s) & $\boldsymbol{0.512 \pm 0.328}$ & $12.01 \pm 0.966$ \\
        %\midrule
        Avg. Time (s) & $\boldsymbol{0.216 \pm 0.064}$ & $8.49 \pm 0.413$ \\
        %\midrule
        Total Length (m) & $\boldsymbol{1.94 \pm 0.742}$ & $3.20 \pm 2.08$ \\
        \bottomrule
    \end{tabular}
\end{table}

Our static-setting experiments were simulated in IsaacSim~\cite{nvidia2025isaacsim} with OmniGibson~\cite{pmlr-v205-li23a}, and our disturbance-setting and scalability experiments were simulated using Drake \cite{drake}. All experiments were run on a local workstation with a 12th Gen Intel i7-12700F CPU and 32~GB RAM.

\subsection{Comparisons to Existing TAMP Methods}
\label{sec:staticanalysis}
In this study, we initially examine our method with ReKep in \textit{static scenarios} without external disturbances, focusing on block-stacking and pick-and-pour tasks. Results are summarized in Tables \ref{tab:blocks} and \ref{tab:pickpour}.

Our method achieves significantly lower computation times and total path lengths, along with higher success rates. While ReKep’s longer computation times stem largely from its optimizer and cost formulations, our method’s superior success rate and solution quality can be attributed to the advantages of the \ac{GoC} structure.

In the block-stacking task, linearizing the \ac{GoC} into a sequence suitable for ReKep introduces substantial idle time for the second robot, since placements must be explicitly interleaved. For the pick-and-pour task, although the \ac{DAG} structure can be linearized while retaining some parallelism in cup handling, ReKep still suffers from single-sequence failure modes. In one case, a failed grasp triggers backtracking of the entire stage, unnecessarily resetting the other robot’s successful grasp. In another, ReKep requires independent path planning for each agent, preventing joint optimization toward a central pouring location. This limitation makes some pouring locations infeasible and leads to misaligned pours. By contrast, our formulation inherently considers agent interactions and potential collisions during optimization, avoiding such failure modes and enabling more robust multi-agent execution.

% We observed in our experiments that GoC-MPC traveled
% a longer total distance ($+1$ m.) than
% ReKep, despite having a lower task completion time.
% One reason relates to its natural multi-agent coordination.
% For example, in some trials both agents moved closer to the subgoal to improve efficiency, after which one left space for the other to complete the task.

\subsection{Robustness to Disturbances}
\label{sec:disturbancesanalysis}

In multi-agent settings, disturbances are particularly challenging: external perturbations may force agent reassignment, delay current and future actions, or alter the set of feasible trajectories. To evaluate the robustness of \ac{GoC-MPC} in such cases, we consider how it adapts to disturbances.

\begin{figure} 
\centering

\subfloat[Initial Grasp]{%
   \includegraphics[width=0.46\columnwidth]{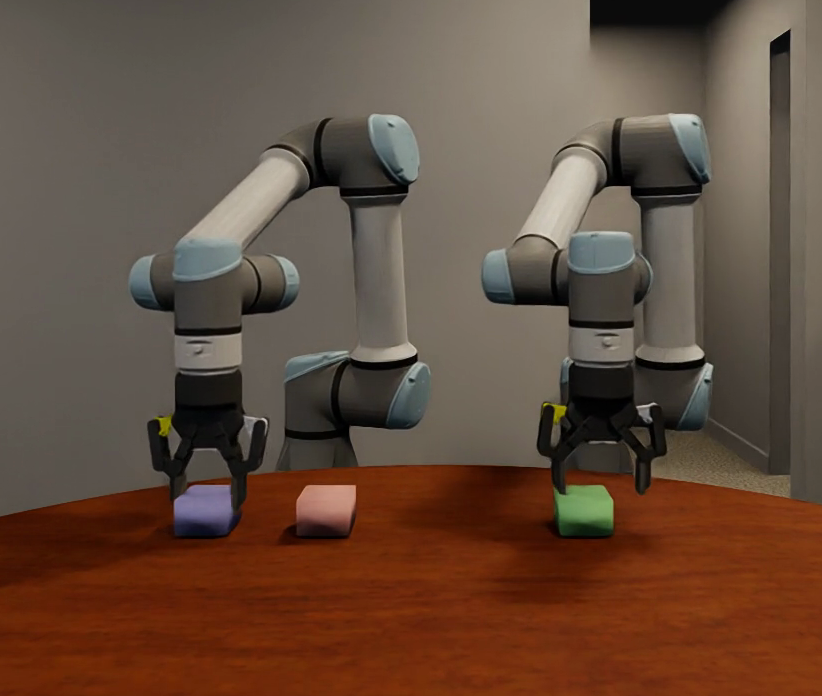}
   \label{fig:grab1}
}
\subfloat[Disturbance]{
   \includegraphics[width=0.46\columnwidth]{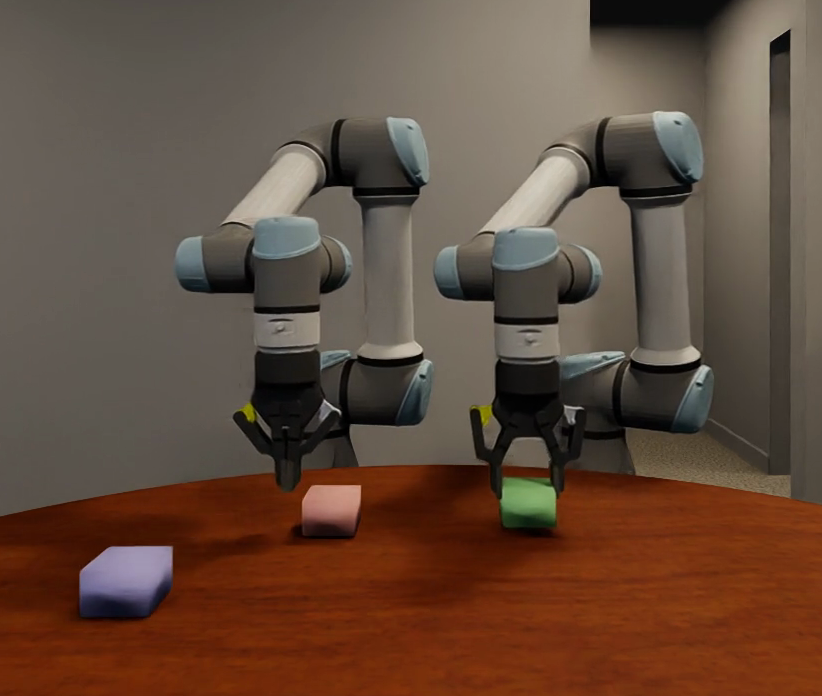}
    \label{fig:disturbed}
} \\
\subfloat[Backtracking]{%
   \includegraphics[width=0.46\columnwidth]{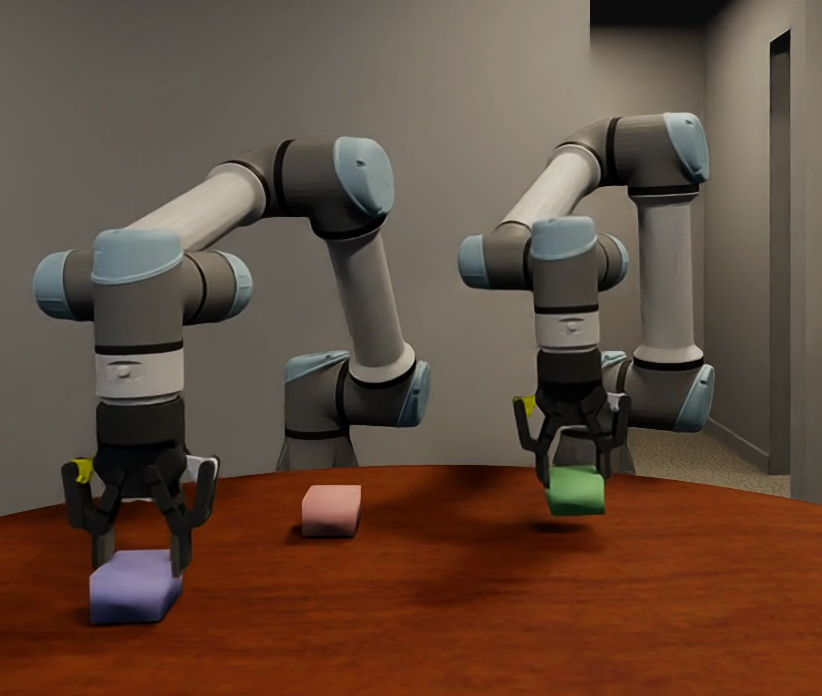}
   \label{fig:backtracked}
}
\subfloat[Task Completion]{
   \includegraphics[width=0.46\columnwidth]{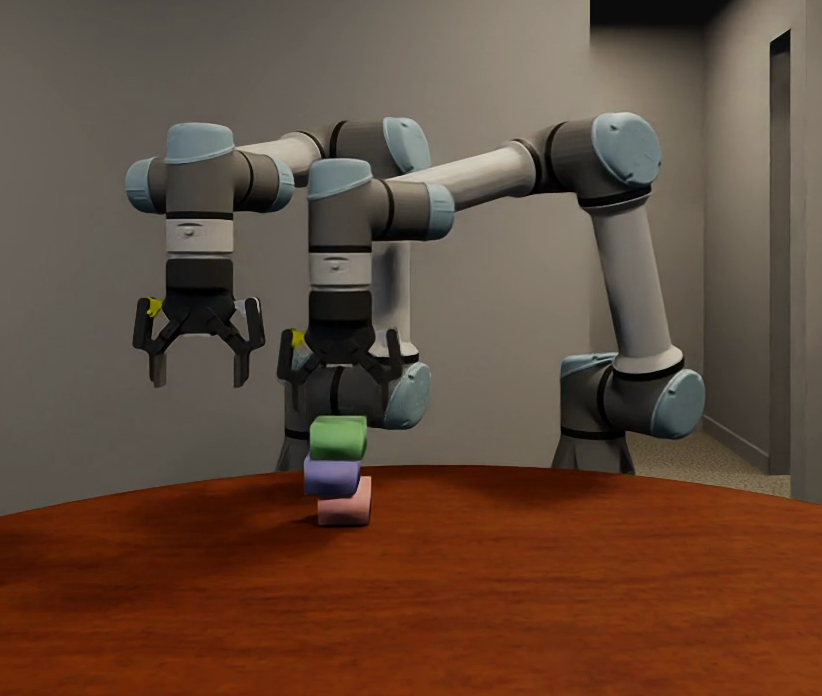}
    \label{fig:finished}
}

\caption{\acp{GoC} allow for agents to independently backtrack to previous stages when a disturbance is applied. In the block stacking task, a disturbance to the block in the left robot's grasp prompts the robot to loop back to picking up the block, and the right robot's path is slowed to the point where it smoothly reaches its original destination after the other robot has finished retrieving the disturbed block.}
\label{fig:independent-backtracking}

\end{figure}

% \begin{table}
% \centering
%     \caption{Success and Computation Complexity under Disturbances}
%     \label{tab:real-world-data-for-disturbances}
%     % \scriptsize
%     % \tiny
%     \setlength{\tabcolsep}{4.0pt}
%     \begin{tabular}{lccc}
%         \toprule
%         \multicolumn{1}{c}{Task} &
%         Max Time (s) $\downarrow$ & Avg. Time (s) $\downarrow$ & Success Rate $\uparrow$ \\
%         \midrule
%         \multirow{1}{*}{Block-Stacking} &
%         {$0.161 \pm 0.093$} & {$0.052 \pm 0.007$} & $10 / 10$ \\
%         \midrule
%         \multirow{1}{*}{Pick-and-Pour} &
%         {} & {} & $11.0 \pm 3.1$ \\
%         \bottomrule
%     \end{tabular}
% \end{table}

\begin{table}[t]
\centering
    %\scriptsize
    % \setlength{\tabcolsep}{4.0pt}
    \caption{Comparison of our approach with ReKep on the block-stacking with external disturbances. While both methods achieved a 100\% success rate, the \textit{agent-independent backtracking} provided by \acp{GoC} saved on average $0.64$ meters of unnecessary movement compared to ReKep, executing 80 $\times$ faster in maximum time.}
    \label{tab:block-stacking-with-disturbances}
    \begin{tabular}{lcc}
        \toprule
        & \textbf{GoC-MPC (Ours)} & \textbf{ReKep} \\
        \midrule
        Success Rate & $\boldsymbol{10/10}$ & $10/10$ \\
        %\midrule
        Max Time (s) & $\boldsymbol{0.161 \pm 0.093}$ & $12.36 \pm 6.180$ \\
        %\midrule
        Avg. Time (s) & $\boldsymbol{0.052 \pm 0.007}$ & $3.29 \pm 0.331$ \\
        %\midrule
        Total Length (m) & $\boldsymbol{6.89 \pm 0.938}$ & $7.53 \pm 0.892$ \\
        \bottomrule
    \end{tabular}
\end{table}

A key property of GoCs is the ability to backtrack when edge constraints are violated. Perturbations may trigger such violations and update the set of active stages—for example, one agent may need to reposition while another continues unaffected. Prior methods do not support coordinated backtracking across agents, whereas our approach does, resulting in shorter paths and better coordination. Fig.~\ref{fig:independent-backtracking} illustrates this behavior in a block-stacking task: when a perturbation occurs, agent~1 backtracks to the pickup stage. Without agent~1’s block, agent~2 cannot complete its subgoal and must delay execution until agent~1 finishes. In this way, GoC-MPC naturally handles interdependent delays and action reordering caused by disturbances.

The time metrics for GoC-MPC were higher in static settings (Table~\ref{tab:blocks}–\ref{tab:pickpour}) than in disturbance experiments (Table~\ref{tab:block-stacking-with-disturbances}). This is because, in static settings, GoC-MPC was initialized with additional stages to enhance robustness, which increased execution time compared to disturbance scenarios.

% Agents can be disabled and We conducted experiments to analyze the overall robustness of our pipeline in completing a variety of real-world tasks, such as block-stacking, transferring liquid in cups, and garment folding. Our results are shown in Table~\ref{tab:real-world-data}.

\subsection{Scalability Analysis}
\label{sec:scalabilityanalysis}

\begin{table}[t]
    \centering
    \caption{Scalability performance of GoC-MPC. The number of agents and blocks in the environment vary across experiments. Mean and standard deviation are reported over 5 randomly initialized trials.}
    \label{tab:scalability}
    \begin{tabular}{@{}cccc}
        \toprule
         (\# Obj, \# Agents) & Success & Avg. Time (s) & Length (m) \\
         \midrule
         {$(5, 2)$} & {5/5} & {$0.319 \pm 0.055$} & {$10.95 \pm 1.35$} \\    
         {$(8, 2)$} & {5/5} & {$0.976 \pm 0.196$} & {$18.53 \pm 3.83$} \\
         {$(11, 2)$} & {4/5} & {$2.462 \pm 0.516$} & {$27.532 \pm 4.445$} \\
         {$(5, 3)$} & {5/5} & {$0.365 \pm 0.103$} & {$12.19 \pm 1.14$} \\
         {$(8, 3)$} & {4/5} & {$1.23 \pm 0.363$} & {$21.50 \pm 4.11$} \\
         {$(11, 3)$} & {4/5} & {$2.650 \pm 1.051$} & {$30.570 \pm 3.783$} \\
         {$(5, 4)$} & {5/5} & {$0.513 \pm 0.184$} & {$14.22 \pm 1.27$} \\
         {$(8, 4)$} & {4/5} & {$1.43 \pm 0.243$} & {$22.70 \pm 2.57$} \\
         {$(11, 4)$} & {5/5} & {$3.594 \pm 1.052$} & {$33.480 \pm 3.998$} \\
         \bottomrule
    \end{tabular}
\end{table}

To assess scalability, we evaluated GoC-MPC's robustness across different numbers of objects
and agents in the scene.
As the number of objects increased, the average time and path lengths
naturally increased, but remained low-enough on average to still support online reactive execution. Success rates also remained high; however, it was occasionally observed with large problems that the waypoints problem could exhaust the allowed number of solver iterations, resulting in some cycles failing to update the planned waypoints. The majority of trials remained successful even with the increased scale.

\subsection{Real-World evaluation}
\label{sec:realworldanalysis}

Finally, we deployed \ac{GoC-MPC} on a dual-UR5e manipulator system to evaluate its performance on real-world bi-manual manipulation tasks. As shown in Fig.~\ref{fig:experiment-settings}, the tasks included transferring liquid between two cups, stacking three blocks, and folding a tablecloth. Demonstration videos are available in the supplementary material. 

Workspace keypoints were tracked from a 30 Hz RGB-D stream using a RealSense D455 camera. For block-stacking and pick-and-pour, block and cup centroids were tracked with SAM2~\cite{ravi2024sam2}, while for tablecloth folding we additionally employed a Kanade–Lucas–Tomasi tracker~\cite{lucas1981iterative, tomasi1991detection} to track corner points. We report maximum and average times per \ac{GoC-MPC} cycle, along with task success rates across five randomly initialized trials (Table~\ref{tab:real-world-data}). Overall, GoC-MPC achieved comparable performance in real-world experiments and simulation, succeeding in nearly all trials. The only failures occurred in two block-stacking trials due to severe block occlusions during execution. These results demonstrate that our approach transfers effectively to physical systems while retaining low computational times and short path lengths.

% Addressing this limitation with a more advanced keypoint tracker is left for future work.

\begin{table}
\centering
    % \scriptsize
    % \tiny
    %\setlength{\tabcolsep}{4.0pt}
    \caption{Real-world Success Rates and Computation Time for our method.}
    \label{tab:real-world-data}
    \begin{tabular}{lccc}
        \toprule
        \multicolumn{1}{c}{Task} &
        Max Time (s) $\downarrow$ & Avg. Time (s) $\downarrow$ & Success $\uparrow$ \\
        \midrule
        \multirow{1}{*}{Block-Stacking} &
        {$0.493 \pm 0.018$} & {$0.093 \pm 0.027$} & $3 / 5$ \\
        %\midrule
        \multirow{1}{*}{Pick-and-Pour} &
        {$0.373 \pm 0.071$} & {$0.065 \pm 0.012$} & $5 / 5$ \\
        %\midrule
        \multirow{1}{*}{Cloth-Folding} &
        {$0.134 \pm 0.009$} & {$0.057 \pm 0.002$} & $5 / 5$ \\
        \bottomrule
    \end{tabular}
\end{table}

\section{FUTURE WORK \& CONCLUSIONS}

In this work, we presented \acp{GoC} and \ac{GoC-MPC}, a novel and effective algorithm for multi-agent reactive \ac{TAMP}. As a generalization of prior work on sequences-of-constraints, \acp{GoC} automatically inherit existing constraint formulations that make them easily implemented for use in the real world without extensive modeling or data requirements. Our experiments additionally show that our approach can formulate and solve \acp{GoC} at almost real-time speeds. However, our method's performance can deteriorate depending on the external state estimation module and the initial plan skeleton. We intend to explore integrating a learned perception module to overcome the former limitation, as well as discrete planning in-the-loop to overcome the latter.

\addtolength{\textheight}{-1cm}   % This command serves to balance the column lengths

%%%%%%%%%%%%%%%%%%%%%%%%%%%%%%%%%%%%%%%%%%%%%%%%%%%%%%%%%%%%%%%%%%%%%%%%%%%%%%%%

%%%%%%%%%%%%%%%%%%%%%%%%%%%%%%%%%%%%%%%%%%%%%%%%%%%%%%%%%%%%%%%%%%%%%%%%%%%%%%%%



% \section*{ACKNOWLEDGMENT}

% The preferred spelling of the word ÒacknowledgmentÓ in America is without an ÒeÓ after the ÒgÓ. Avoid the stilted expression, ÒOne of us (R. B. G.) thanks . . .Ó  Instead, try ÒR. B. G. thanksÓ. Put sponsor acknowledgments in the unnumbered footnote on the first page.

\begin{acronym}[TDMA]
\acro{LVM}{Large Vision Model}
\acro{VLM}{Vision Language Model}
\acro{DSL}{Domain Specific Language}
\acro{TAMP}{Task and Motion Planning}
\acro{MA-TAMP}{Multi-Agent Task and Motion Planning}
\acro{ADMM}{Alternating Direction Method of Multipliers}
\acro{C-ADMM}{Consensus Alternating Direction Method of Multipliers}
\acro{MPC}{Model Predictive Control}
\acro{D-MPC}{Distributed Model Predictive Control}
\acro{GoC-MPC}{Graph-of-Constraints Model Predictive Control}
\acro{GoC}{Graph-of-Constraints}
\acrodefplural{GoC}{Graphs-of-Constraints}
\acro{ReKep}{Relational Keypoint constraint}
\acro{DAG}{Directed Acyclic Graph}
\end{acronym}

%%%%%%%%%%%%%%%%%%%%%%%%%%%%%%%%%%%%%%%%%%%%%%%%%%%%%%%%%%%%%%%%%%%%%%%%%%%%%%%%

% References are important to the reader; therefore, each citation must be complete and correct. If at all possible, references should be commonly available publications.

\bibliographystyle{IEEEtran}
\bibliography{IEEEabrv,references}

\end{document}